\documentclass[conference]{IEEEtran}
\IEEEoverridecommandlockouts
\usepackage{cite}
\usepackage{amsthm}
\theoremstyle{definition}
\newtheorem{definition}{Definition}
\usepackage{amsmath,amssymb,amsfonts}
\usepackage{threeparttable}
\usepackage{graphicx}
\usepackage{textcomp}
\usepackage{algorithm}
\usepackage{xcolor}
\usepackage{algpseudocode}
\usepackage{enumitem}   
\usepackage{subfigure}
\begin{document}

\title{Zadeh's Type-2 Fuzzy Logic Systems: Precision and High-Quality Prediction Intervals \\
\thanks{This work was supported by MathWorks\textsuperscript{\textregistered} in part by a Research Grant awarded to T. Kumbasar. Any opinions, findings, conclusions, or recommendations expressed in this paper are those of the authors and do not necessarily reflect the views of the MathWorks, Inc.}

}


 \author{\IEEEauthorblockN{Yusuf Güven}
 \IEEEauthorblockA{\textit{Control and Automation Eng. Dept.} \\
 \textit{Istanbul Technical University}\\
 Istanbul, Türkiye \\
 guveny18@itu.edu.tr}
 \and
 \IEEEauthorblockN{Ata Köklü}
 \IEEEauthorblockA{\textit{Control and Automation Eng. Dept.} \\
 \textit{Istanbul Technical University}\\
Istanbul, Türkiye \\
 koklu18@itu.edu.tr}
 \and
 \IEEEauthorblockN{Tufan Kumbasar}
 \IEEEauthorblockA{\textit{Control and Automation Eng. Dept.} \\
 \textit{Istanbul Technical University}\\
 Istanbul, Türkiye \\
 kumbasart@itu.edu.tr}}

\maketitle

\begin{abstract}
General Type-2 (GT2) Fuzzy Logic Systems (FLSs) are perfect candidates to quantify uncertainty, which is crucial for informed decisions in high-risk tasks, as they are powerful tools in representing uncertainty. In this paper, we travel back in time to provide a new look at GT2-FLSs by adopting Zadeh's (Z) GT2 Fuzzy Set (FS) definition, intending to learn GT2-FLSs that are capable of achieving reliable High-Quality Prediction Intervals (HQ-PI) alongside precision. By integrating Z-GT2-FS with the \(\alpha\)-plane representation, we show that the design flexibility of GT2-FLS is increased as it takes away the dependency of the secondary membership function from the primary membership function. After detailing the construction of Z-GT2-FLSs, we provide solutions to challenges while learning from high-dimensional data: the curse of dimensionality, and integrating Deep Learning (DL) optimizers. We develop a DL framework for learning dual-focused Z-GT2-FLSs with high performances. Our study includes statistical analyses, highlighting that the Z-GT2-FLS not only exhibits high-precision performance but also produces HQ-PIs in comparison to its GT2 and IT2 fuzzy counterparts which have more learnable parameters. The results show that the Z-GT2-FLS has a huge potential in uncertainty quantification.
\end{abstract}


\section{Introduction}

Deep Learning (DL) has achieved significant success in various applications. Yet, the main focus is on accuracy which is inadequate in high-risk tasks. Hence, employing Uncertainty Quantification (UQ) is essential \cite{abdar2021review, papadopoulos2011regression, pearce, PINAW}. For UQ, General Type-2 (GT2) Fuzzy Logic Systems (FLSs), and also its simplified version Interval Type-2 (IT2), are perfect candidates as they are powerful tools in representing uncertainty thanks to their Membership Functions (MFs) defined with GT2 Fuzzy Sets (FSs) \cite{sakalli, almaraash2023life, pekaslan2019leveraging,mendel2018comparing}. While Zadeh (Z) initially defined GT2-FSs \cite{ZADEH-1, ZADEH-2}, the fundamental research studies started with the GT2-FS definition of Mendel \& John (MJ) \cite{Mendel-2}. Yet, the wide deployment of MJ-GT2-FSs was catalyzed by the zSlices/$\alpha$-plane representation \cite{zslice, mendel_kitap}, driven by two key aspects: (1) the shape parameterization of Secondary MF (SMF) of GT2-FS based on its Primary MF (PMF), and (2) the equivalence of GT2-FLSs with a collection of $\alpha$-plane associated IT2-FLSs. 

Rather than exploding the huge potential of IT2/GT2 FSs in UQ, the mainstream learning methods for IT2/GT2 FLSs focus predominantly on accuracy improvements\cite{shihabudheen2018recent, zheng2021fusion, wiktorowicz2023t2rfis,tavoosi2021review,han2021type}. Recently, there has been a shift in focus, and the IT2 and GT2 FSs have been exploited as an opportunity for UQ. In \cite{beke}, a DL framework is presented via a composite loss that utilizes the Type-Reduced Set (TRS) and output of IT2-FLSs to learn predictors capable of generating Prediction Intervals (PIs). Whereas for MJ-GT2-FLS \cite{avci}, a composite loss is proposed that associates the support of SMFs (i.e. PMFs) for UQ while the shape for accurate pointwise estimation.

In this paper, we present a fresh look at GT2-FLSs through the perspective of Zadeh’s GT2-FS definition, aiming to develop high-performing GT2-FLSs with UQ capability. We show that through the deployment of Z-GT2-FS rather than MJ-GT2-FS, the dependence of the SMF's shape on interval-valued primary membership grade is eliminated and thus provides more design flexibility. In this context, we first provide the mathematical foundations of the SMF and PMF which are both defined with Type-1 FSs. Then, we integrate the $\alpha$-plane method into the Z-GT2-FS to define the output of the Z-GT2-FLS. In this context, to define $\alpha$-planes, we define the $\alpha$-cuts of SMF and then extract equivalent LMF and UMF corresponding to $\alpha$-planes of the Z-GT2-FS to directly use the output formulation defined via $\alpha$-plane representation. 

To address the challenges of learning from high-dimensional data, we propose an approach to mitigate the impact of the curse of dimensionality on the Z-GT2-FLS learning process. The method involves adjusting PMF in proportion to the input dimension. Also, to satisfy the constraints arising from the definitions of FSs during learning, we present parameterization tricks so that learning GT2-FLSs with built-in unconstrained DL optimizers is feasible. 

Within the scope of this study, we present a DL framework to learn a dual-focused Z-GT2-FLS that not only yields accurate point-wise predictions but also excels in generating High-Quality (HQ) PI. To show the superiority of the proposed approach alongside Z-GT2-FLS, we analyze and compare the learning performance of the Z-GT2-FLS on high-dimensional datasets. This comparison includes its MJ-GT2 and IT2 fuzzy counterparts that have a higher number of Learnable Parameters (LPs). The statistical performance analysis demonstrates the capability of learning Z-GT2-FLS as a viable solution for achieving reliable HQ-PI with a high degree of precision.


\section{GT2-FLSs: A Brief Overview} \label{GT2-FLS}
The GT2-FLS is formulated for an input vector $\mathbf{x}=$ $\left(x_{1}, x_{2}, \ldots, x_{M}\right)^{\mathrm{T}}$  and a single output $y$. The rule base is composed of $P$ rules $(p=1,2, \ldots, P)$ that is defined as:
\begin{equation}\label{rule}
R_{p}: \text{If } x_{1} \text{ is } \tilde{A}_{p, 1} \text{ and} \ldots x_{M} \text{ is }\tilde{A}_{p, M} \text{ Then } y \text{ is } y_p 
\end{equation}
where $y_{p}$ represents the consequent MFs that are defined as:
\begin{equation}
    y_{p} = \sum_{m=1}^{M} a_{p, m} x_{m} + a_{p, 0}
    \label{eq1}
\end{equation}
The antecedent MFs are defined with GT2-FSs $\tilde{A}_{p, m}$ that are defined as a collection of $\alpha$-planes $\left(\alpha_{k}\right)$ as follows:
\begin{equation}
    \tilde{A}_{p, m}=\bigcup_{\alpha_{k} \in[0,1]} \tilde{A}_{p, m}^{\alpha_{k}}
    \label{eq2}
\end{equation}
where $\tilde{A}_{p, m}^{\alpha_{k}}$ is the $\alpha$-plane of $\tilde{A}_{p, m}$ associated with $\alpha_{k} \in[0,1]$. When $\alpha_{k}$ is distributed uniformly, we express it as $\alpha_{k} = k / K$ for $k$ ranging from 0 to $K$. Thus, there are a total of $K$ + 1 $\alpha$-planes \cite{mendel_kitap}. This representation allows defining the output of the GT2-FLSs as follows:

\begin{equation} \label{alphap}  y=\frac{\sum_{k=0}^{K} y^{\alpha_{k}}(\boldsymbol{x}) \alpha_{k}}{\sum_{k=0}^{K} \alpha_{k}}
\end{equation}
where $\mathrm{y}^{\alpha_{k}}(\boldsymbol{x})$ is the output of an IT2-FLS associated with an $\alpha$-plane $\alpha_{k}\left(\alpha_{k}\right.$-IT2-FLS) that is defined as:
\begin{equation}
    y^{\alpha_{k}}(\boldsymbol{x})=
(\underline{y}^{\alpha_{k}}(\boldsymbol{x})+\overline{y}^{\alpha_{k}}(\boldsymbol{x}))/2
    \label{eq4}
\end{equation}
Here, $[\underline{y}^{\alpha_{k}}, \overline{y}^{\alpha_{k}}]$ is type reduced set of $\alpha_{k}$-IT2-FLS: 
\begin{equation}
\begin{split} 
\underline{y}^{\alpha_{k}}(\boldsymbol{x}) = \frac{\sum_{p=1}^{L} \underline{f}_{p}^{\alpha_{k}}(\boldsymbol{x}) {y}_{{p}} + \sum_{p=L+1}^{P} \overline{f}_{p}^{\alpha_{k}}(\boldsymbol{x}) {y}_{{p}}}{\sum_{p=1}^{L} \underline{f}_{p}^{\alpha_{k}}(\boldsymbol{x}) + \sum_{p=L+1}^{P} \overline{f}_{p}^{\alpha_{k}}(\boldsymbol{x})} \\
\overline{y}^{\alpha_{k}}(\boldsymbol{x}) = \frac{\sum_{p=1}^{R} \underline{f}_{p}^{\alpha_{k}}(\boldsymbol{x}) {y}_{{p}} + \sum_{p=R+1}^{P} \overline{f}_{p}^{\alpha_{k}}(\boldsymbol{x}) {y}_{{p}}}{\sum_{p=1}^{R} \underline{f}_{p}^{\alpha_{k}}(\boldsymbol{x}) + \sum_{p=R+1}^{P} \overline{f}_{p}^{\alpha_{k}}(\boldsymbol{x})}
\label{eq6}
\end{split}
\end{equation}
where $L, R$ are the switching points of the Karnik-Mendel algorithm \cite{mendel_kitap}.  $\underline{f}_{p}^{\alpha_{k}}(\boldsymbol{x})$ and $\overline{f}_{p}^{\alpha_{k}}(\boldsymbol{x})$ are the lower and upper rule firing of the $p^{th}$ rule and are defined as:
\begin{equation}
\begin{split} 
\underline{f}_{p}^{\alpha_{k}}(\boldsymbol{x}) = \underline{\mu}_{\tilde{A}_{p, 1}^{\alpha_{k}}}\left(x_{1}\right) \cap \underline{\mu}_{\tilde{A}_{p, 2}^{\alpha_{k}}}\left(x_{2}\right) \cap \ldots \cap \underline{\mu}_{\tilde{A}_{p, M}^{\alpha_{k}}}\left(x_{M}\right) \\
\overline{f}_{p}^{\alpha_{k}}(\boldsymbol{x}) = \overline{\mu}_{\tilde{A}_{p, 1}^{\alpha_{k}}}\left(x_{1}\right) \cap \overline{\mu}_{\tilde{A}_{p, 2}^{\alpha_{k}}}\left(x_{2}\right) \cap \ldots \cap \overline{\mu}_{\tilde{A}_{p, M}^{\alpha_{k}}}\left(x_{M}\right)
\label{upper_firing}
\end{split}
\end{equation}
Here, $\cap$ denotes the t-norm operator which can be defined with the product or min operator \cite{mendel_kitap}.

\section{Learning GT2-FLSs: Flexibility and DL}\label{Section3}
In this section, we present all the details on how to represent and learn high-performing GT2-FLSs for high-dimensional data via DL optimizers efficiently. 
\subsection{MJ-GT2-FLS: Representation and Potential Issues} \label{Batu}
To define $\tilde{A}_{p, m}$, the most widely used GT2-FS representation is the one of Mendel \& John \cite{Mendel-2}.
\begin{definition} \label{mendel} 
A GT2-FS $\tilde{A}$ is characterized by a type-2 MF $\left(x, \mu_{\tilde{A}}(x, u)\right)$, where $x \in X$ and $u \in J_x \subseteq[0,1]$, i.e., 
\begin{equation}
\tilde{A}=\left\{\left(x, \mu_{\tilde{A}}(x, u)\right) \mid x \in X, u \in J_x \subseteq[0,1]\right\}
\end{equation}
in which $0 \le \mu_{\tilde{A}}(x, u) \le 1$ \cite{Mendel-2}.
\end{definition}

\begin{figure*}[hbpt]
        \centering
        \subfigure[MJ-GT2-FS]
        {
        \includegraphics[width=0.4\textwidth]
                    {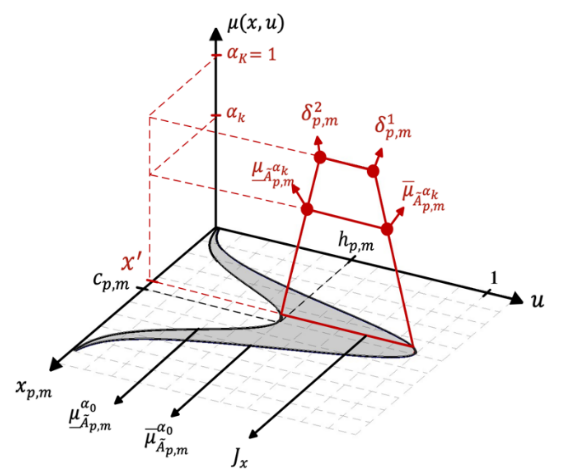}
                    \label{MJ-GT2-FS}
        }
        \subfigure[Z-GT2-FS]
        {
        \includegraphics[width=0.41\textwidth]
                    {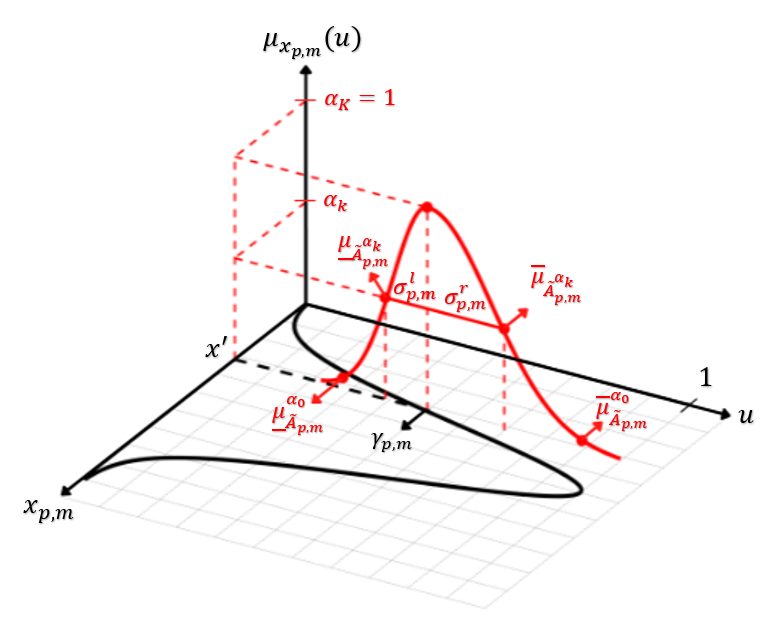}
                    \label{Z-GT2-FS}
        
        }
          \caption{Illustrations of a GT2-FS with an $\alpha$ - plane}
          \label{GT2-FSs}
\end{figure*}

We first parameterize a PMF, i.e. $\tilde{A}_{p, m}^{\alpha_{0}}$, by defining the following Upper MF (UMF) $\bar{\mu}_{\tilde{A}_{p, m}}^{\alpha_{0}}\left(x_{m}\right)$ and Lower MF (LMF) $\underline{\mu}_{\tilde{A}_{p, m}}^{\alpha_{0}}\left(x_{m}\right)$:
\begin{equation}
\begin{split} 
\overline{\mu}_{\tilde{A}_{p, m}}^{\alpha_{0}}\left(x_{m}\right)=\exp \left(-\left(x_{m}-c_{p, m}\right)^{2} / 2 \overline{\sigma}_{p, m}^{2}\right)  \\
\underline{\mu}_{\tilde{A}_{p, m}}^{\alpha_{0}}\left(x_{m}\right)=h_{p, m} \exp \left(-\left(x_{m}-c_{p, m}\right)^{2} / 2 \underline{\sigma}_{p, m}^{2}\right)
\label{lower_upper_mendel}
\end{split}
\end{equation}
where $c_{p, m}$ is the center, $\tilde{\sigma}_{p, m}=[\underline{\sigma}_{p, m},\overline{\sigma}_{p, m}]$ is the standard deviation while $h_{p, m}$ defines the height of the LMF $\forall p, m$. As depicted in Fig. \ref{MJ-GT2-FS}, the Footprint Of Uncertainty $\left(J_{x}\right)$ or the support of the SMF is defined with $\tilde{\sigma}_{p, m}$ and $h_{p, m}$.
Based on the PMF, the following parameterized SMF is used that is defined via the UMF and LMF of $\tilde{A}_{p, m}^{\alpha_{k}} (k \neq 0)$ \cite{sakalli}:
\begin{equation}
\begin{split} 
    \underline{\mu}_{\tilde{A}_{p, m}^{\alpha_{k}}}=\underline{\mu}_{\tilde{A}_{p, m}}^{\alpha_{0}}+\alpha_{k}\left(\overline{\mu}_{\tilde{A}_{p, m}}^{\alpha_{0}}-\underline{\mu}_{\tilde{A}_{p, m}}^{\alpha_{0}}\right) \delta_{p, m}^{1} \\
    \overline{\mu}_{\tilde{A}_{p, m}^{\alpha_{k}}}=\overline{\mu}_{\tilde{A}_{p, m}}^{\alpha_{0}}-\alpha_{k}\left(\overline{\mu}_{\tilde{A}_{p, m}}^{\alpha_{0}}-\underline{\mu}_{\tilde{A}_{p, m}}^{\alpha_{0}}\right)\left(1-\delta_{p, m}^{2}\right)
\end{split}
\label{newmulowermuupper}
\end{equation}
Here, $\left\{\delta_{p, m}^{1}, \delta_{p, m}^{2}: \delta_{p, m}^{1} \geq \delta_{p, m}^{2}\right\} \in[0,1], \forall p, m$ are parameters that define the shape of the SMFs as shown in Fig. \ref{MJ-GT2-FS}.  

In a recent study \cite{avci}, a DL-based learning method for MJ-GT2-FLS, which is based on IT2-FLS one in \cite{beke}, is presented. We identified the following two problems: 
\begin{enumerate} [label=(\roman*)]
  \item \textbf{Flexibility:} The drawback of the method lies in its insistence on the explicit parameterization of SMFs with respect to the PMFs prior to training, i.e. the implementation of Definition \ref{mendel} via \eqref{newmulowermuupper}. While providing structure, this might hinder the GT2-FLS's learning capacity as the learning performance depends on how the shapes of the UMF and LMF in \eqref{lower_upper_mendel} (i.e. the Footprint of Uncertainty size) are defined. 
  \item \textbf{Curse of dimensionality:} They handled this problem by setting the t-norm operator $\cap$ in \eqref{upper_firing} w.r.t the data size ($N$) and dimension ($M$). They suggested using the product operator for low dimensional input vector spaces while the min one for high dimensional ones based on their exhaustive comparative results.  
\end{enumerate}


\subsection{Z-GT2-FLS: Representation and Solutions} \label{oursversion}

In this section, we travel back in time and define the GT2-FSs with the one of Zadeh to provide flexibility during the learning and also provide a solution to the curse of dimensionality problem. 



\subsubsection{Zadeh's representation of a GT2-FLSs}
Let us start with the definition of Zadeh \cite{ZADEH-1},\cite{ZADEH-2} to define GT2-FSs. 
\begin{definition}  \label{zadeh}
A GT2-FS $\tilde{A}$ on $X$ is a mapping $\tilde{A}: X \rightarrow \text{FS}\left([0, 1]\right)$ \cite{ZADEH-1}.
\end{definition}
\noindent Alternatively, we can state that $\tilde{A}_{p, m}$ is mapping $\tilde{A}_{p, m}: X \rightarrow [0, 1]^{[0,1]}$. The equivalence of Definition \ref{mendel} and \ref{zadeh} is given \cite{bustince}. 

In this study, we adopted Definition \ref{zadeh} and integrated it with $\alpha$-plane representation to define the output of Z-GT2-FLS via \eqref{alphap}. As shown in Fig. \ref{Z-GT2-FS}, we represent the PMF with a Type-1 FS $A_{p, m}$ that is defined as follows:
\begin{equation}
    {\mu}_{{A}_{p, m}}(x_m) = \exp \left (-\left (x_{m} - c_{p, m}\right)^{2} / 2\sigma_{p, m}^{2}\right )
    \label{pmf}
\end{equation}
while the SMF with a two-sided Gaussian MF is as follows:
\begin{equation}
 {\mu_{x_{p,m}}(u)}= \begin{cases}
    \begin{aligned}
        &\exp \left( -\frac{(u - {\gamma_{{p, m}}})^{2}}{2(\sigma_{p, m}^l)^2} \right), \text{if } {u} \leq {\gamma_{{p, m}}} \\
        &\exp \left( -\frac{(u - {\gamma_{{p, m}}})^{2}}{2(\sigma_{p, m}^r)^2} \right), \text{if } {u} \geq {\gamma_{{p, m}}}
    \end{aligned}
 \end{cases}
 \label{smf}
\end{equation}
where $\sigma^l_{p, m}$ and $\sigma^r_{p, m}$ are the left and right standard deviations and $\gamma_{{p, m}}$ is the center, which define the shape and support of the SMF. As shown in Fig. \ref{Z-GT2-FS}, we set $\gamma_{{p, m}}={\mu}_{{A}_{p, m}}(x_m)$. 

Now, to extract the $\alpha$-planes of the Z-GT2-FS ($\tilde{A}_{p, m}^{\alpha_k}$), we define the $\alpha$-cuts of $\mu_{x_{p, m}}(u)$. We first rewrite \eqref{smf} as follows: 
\begin{equation} \label{eq13}
         -\sqrt{-2 \ln \left(\alpha_k\right)} \sigma^l_{p, m} = \left (u - {\gamma_{{p, m}}}\right), \text{if } {u} \leq {\gamma_{{p, m}}}
\end{equation}
\begin{equation} \label{eq14}
           \sqrt{-2 \ln \left(\alpha_k\right)} \sigma^r_{p, m} = \left (u- {\gamma_{{p, m}}}\right), \text{if } {u} \geq {\gamma_{{p, m}}}
\end{equation}
Then, by inserting $\underline{\mu}_{\tilde{A}_{p, m}^{\alpha_{k}}}$ into \eqref{eq13} while $\overline{\mu}_{\tilde{A}_{p, m}^{\alpha_{k}}}$ into \eqref{eq14} as $u$, we can extract the UMF and LMF of $\tilde{A}_{p, m}^{\alpha_{k}} (k \neq 0)$: 
\begin{equation}
\begin{split}
\underline{\mu}_{\tilde{A}_{p, m}^{\alpha_{k}}}(x_m)={\mu}_{{A}_{p, m}}(x_m)-\sqrt{-2 \ln \left(\alpha_k\right)} \sigma^l_{p, m} \\    \overline{\mu}_{\tilde{A}_{p, m}^{\alpha_{k}}}(x_m)={\mu}_{{A}_{p, m}}(x_m)+\sqrt{-2 \ln \left(\alpha_k\right)} \sigma^r_{p, m}
    \label{muupper}
    \end{split}
\end{equation}
Now, by using \eqref{muupper}, we can define the output of Z-GT2-FLS via $\alpha$-plane representation as given in \eqref{alphap}. The only problem with this implementation is due to the domain space of $\ln(\cdot)$, which spans $(0, \infty]$, and thus $\alpha_{0}=0$ is not included. Thus, we associate the $\alpha_{0}$-plane with $\alpha_0 \triangleq 0.01$. Our motivation for this setting is grounded in the consideration that $\sqrt{-2 \ln \left(\alpha_0\right)}$ $\approx$ 3. Thus, $\underline{\mu}_{\tilde{A}_{p, m}}^{\alpha_0}$ and $\overline{\mu}_{\tilde{A}_{p, m}}^{\alpha_0}$ are defined explicitly as follows:
\begin{equation}
\begin{split}
    \underline{\mu}_{\tilde{A}_{p, m}}^{\alpha_0}(x_m) \approx {\mu}_{{A}_{p, m}}(x_m) - 3\sigma^l_{p, m} \\
    \overline{\mu}_{\tilde{A}_{p, m}}^{\alpha_0}(x_m) \approx {\mu}_{{A}_{p, m}}(x_m) + 3\sigma^r_{p, m}
    \label{maxvalue}
\end{split}
\end{equation}
Thus, we can claim that $\underline{\mu}_{\tilde{A}_{p, m}}^{\alpha_0}$ and  $\overline{\mu}_{\tilde{A}_{p, m}}^{\alpha_0}$ remain within a range of three standard deviations from ${\gamma_{{p, m}}}$ approximately.

\subsubsection{Curse of dimensionality}
Unlike \cite{avci}, our study adopts a consistent approach by using the product operator as $\cap$ in (\ref{upper_firing}) regardless of the data size and dimension. To avoid the general problem of $[\underline{f}_{p}^{\alpha_{k}}(\boldsymbol{x}), 
\overline{f}_{p}^{\alpha_{k}}(\boldsymbol{x})] \rightarrow 0$ in high-dimensional datasets (i.e. rule firing problem), we propose a method like the HTSK \cite{HTSK} that scales ${\mu}_{{A}_{p, m}}$ w.r.t $M$ as follows:
\begin{equation}
{{\mu}^*_{{A}_{p, m}}}=({\mu}_{{A}_{p, m}})^{1/M} 
\label{mu_new}
\end{equation}
We can reformulate (\ref{mu_new}) in the following explicit standard form of a Gaussian MF to represent the PMF:
\begin{equation}
    {\mu}^*_{{A}_{p, m}}(x_m) = \exp \left (-\left (x_{m} - c_{p, m}\right)^{2} / (2(\sqrt{M}\sigma_{p, m})^{2})\right )
    \label{newpmf}
\end{equation}
which is mathematically equivalent to HTSK \cite{HTSK} as shown in \cite{ata}. When compared to \eqref{pmf}, we can observe that $\sigma_{p, m}$ of the PMF is scaled with $\sqrt{M}$. In this way, the learning of GT2-FLS is not substantially affected by the increase in feature dimensionality.

\subsection{Learnable Parameter Sets}

The LP sets of the presented GT2-FLSs $(\boldsymbol{\theta})$ consist of the LPs of antecedent MF $(\boldsymbol{\theta}_{A})$ and those of the consequent MFs $(\boldsymbol{\theta}_{C})$. For both the MJ-GT2-FLS and Z-GT2-FLS, we define the identical $\boldsymbol{\theta}_{C}$ as $\boldsymbol{\theta}_{\mathrm{C}} = \{ \boldsymbol{a}, \boldsymbol{a}_{0} \}$, with $\boldsymbol{a} = (\boldsymbol{a}_{1,1}, \ldots, \boldsymbol{a}_{P,M})^{T} \in \mathbb{R}^{P \times M}$,  $\boldsymbol{a}_{0} = (a_{1,0}, \ldots, a_{P,0})^{T} \in \mathbb{R}^{P \times 1}$. The only difference between them lies in $\boldsymbol{\theta}_{\mathrm{A}} = \{\boldsymbol{\theta}_{\mathrm{AP}},\boldsymbol{\theta}_{\mathrm{AS}}\}$, i.e., Definition \ref{mendel} vs. Definition \ref{zadeh}.
\begin{itemize}
    \item For MJ-GT2-FLS: $\boldsymbol{\theta}_{\mathrm{AP}}=\{\boldsymbol{c}, \boldsymbol{\sigma}, \boldsymbol{h}\}$  and $\boldsymbol{\theta}_{\mathrm{AS}}=\{ \boldsymbol{\delta}^{(1)}, \boldsymbol{\delta}^{(2)}\}$ with  $\boldsymbol{c} = (c_{1,1}, \ldots, c_{P,M})^{T} \in \mathbb{R}^{P \times M}$, $\boldsymbol{\sigma} = (\sigma_{1,1}, \ldots,\sigma_{P,M})^{T} \in \mathbb{R}^{P \times M}$, $\boldsymbol{h} = (h_{1,1}, \ldots, h_{P,M})^{T} \in \mathbb{R}^{P \times M}$, $\boldsymbol{\delta}^{(1)}=(\delta_{1}^{(1)}, \ldots, \delta_{M}^{(1)})^{T} \in \mathbb{R}^{1 \times M}$, and $\boldsymbol{\delta}^{(2)} = (\delta_{1}^{(2)}, \ldots, \delta_{M}^{(2)})^{T} \in \mathbb{R}^{1 \times M}$. In \cite{avci}, they set $\delta_{p, m}^{1}=\delta_{m}^{1}$ and $\delta_{p, m}^{2}=$ $\delta_{m}^{2},  \forall p$
    \item For Z-GT2-FLS,   $\boldsymbol{\theta}_{\mathrm{AP}}=\{\boldsymbol{c}, \boldsymbol{\sigma}\}$  and $\boldsymbol{\theta}_{\mathrm{AS}}=\{ \boldsymbol{\sigma^l}, \boldsymbol{\sigma^r}\}$ with $\boldsymbol{\sigma}^{(l)}=(\sigma_{1}^{(l)}, \ldots, \sigma_{M}^{(l)})^{T} \in \mathbb{R}^{1 \times M}$, and $\boldsymbol{\sigma}^{r} = (\sigma_{1}^{(r)}, \ldots, \sigma_{M}^{(r)})^{T} \in \mathbb{R}^{1 \times M}$. For the sake of simplicity, we set $\sigma^l_{p, m}=\sigma^l_{m}$ and $\sigma^r_{p, m}=\sigma^r_{m}, \forall p$.
\end{itemize}

To sum up, MJ-GT2-FLS has a total of $(3P + 2)M + P(M + 1)$, while the Z-GT2-FLS involves $(2P + 2)M + P(M + 1)$ LPs. Despite the added complexity of using a Gaussian SMF, the Z-GT2-FLS has $PM$ fewer LPs compared to MJ-GT2-FLS.

\subsection{Parameterization Tricks for DL Optimizers}\label{AA}

The learning problem definition of GT2-FLSs is defined with constraints $\boldsymbol{\theta} \in \boldsymbol{C}$ that are arising from the definitions of FSs and FLSs \cite{beke}. Given that DL optimizers are unconstrained optimization techniques, we introduce parametrization tricks to transform $\boldsymbol{\theta}$ to an unbounded search space. We present only the tricks for $\boldsymbol{\sigma^l}, \boldsymbol{\sigma^r} \in \boldsymbol{\theta}_{{A}}$. For the tricks of the remaining parameters, please check \cite{beke, avci}. 

For $\boldsymbol{\sigma^l}, \boldsymbol{\sigma^r} \in \boldsymbol{\theta}_{{A}}$, we must ensure that the learned $\tilde{A}_{p, m}^{\alpha_{k}}$ adhere to the conditions of GT2-FSs, specifically $0 \leq \underline{\mu}_{\tilde{A}_{p, m}^{\alpha_{k}}}(x_{m}) \leq \bar{\mu}_{\tilde{A}_{p, m}^{\alpha_{k}}}(x_{m}) \leq 1,  \forall p, m$. 
It is important to highlight that, as per the definitions in (\ref{muupper}), we inherently ensure $\underline{\mu}_{\tilde{A}_{p, m}^{\alpha_{k}}}(x_{m}) \leq \overline{\mu}_{\tilde{A}_{p, m}^{\alpha_{k}}}(x_{m}), \forall p, m$. 

\begin{itemize}
    \item For $\boldsymbol{\sigma^l}$, we address the constraint $0 \leq \sigma^l_{m} \leq {\gamma_{p, m}} / \sqrt{-2 \ln(0.01)} $ via:
    \begin{equation}
        \sigma^l_{m} = {\gamma_{p, m}}/\sqrt{-2 \ln(0.01)}\operatorname{sig}({\widehat{\sigma}^l_m})
        \label{trick1}
    \end{equation}
    \item For $\boldsymbol{\sigma^r}$, where the constraint is $0 \leq \sigma^r_{m} \leq \left(1- {\gamma_{p, m}}\right) / \sqrt{-2 \ln(0.01)}$, we do the following trick:
    \begin{equation}
        \sigma^r_{m} = \left(1-{\gamma_{p, m}}\right)/\sqrt{-2 \ln(0.01)}\operatorname{sig}({\widehat{\sigma}^r_m})
        \label{trick2}
    \end{equation}

where $\operatorname{sig}(\cdot)$ is the sigmoid function that provides the generation of unbounded optimization variables $\{{\widehat{\sigma}^l_m},{\widehat{\sigma}^r_m}\}$.
\end{itemize}
Here, the utilization of $1/\sqrt{-2 \ln(0.01)}$ is motivated by the objective to ensure $[\underline{\mu}_{\tilde{A}_{p, m}^{\alpha_{k}}}(x_{m}), \overline{\mu}_{\tilde{A}_{p, m}^{\alpha_{k}}}(x_{m})] \in [0, 1], \forall p, m$.

The presented parameterization tricks alongside the ones in \cite{beke} allow learning the GT2-FLSs via DL optimizers and automatic differentiation methods provided within DL frameworks such as Matlab and PyTorch.

\section{Dual-Focused GT2-FLS: Accuracy \& PI} \label{Dual-Focused}

We introduce a DL framework designed to enable the learning of GT2-FLSs that not only yield accurate point-wise predictions but also excel in generating HQ-PI. {Algorithm~1} provides the training steps of dual-focused GT2-FLS for a dataset $\left\{\boldsymbol{x}_{n}, y_{n}\right \}_{n=1}^{N}$, where $\boldsymbol{x}_{\boldsymbol{n}}=\left(x_{n, 1}, \ldots, x_{n, M}\right)^{T}$.

As we aim to learn a dual-focused GT2-FLS, we defined the following loss to be minimized by a DL optimizer \cite{beke}:  
\begin{equation} \label{loss}
    \min _{\boldsymbol{\theta} \in \mathcal{C}} L=\frac{1}{N}\sum_{n=1}^{N}\left(L_{R}\left(x_n, y_n\right) + \ell\left(x_n, y_n, \underline{\tau}, \overline{\tau}\right)\right)
\end{equation}
 Here, $\boldsymbol{C}$ represents the constraints that can be eliminated as described in Section III.D. The loss function has an accuracy-focused part $L_{R}(\cdot)$ and an uncertainty-focused part $\ell(\cdot)$. 

For the accuracy-focused part $L_{R}(\cdot)$, we use the following  empirical risk function:
\begin{equation}
    L_{R}(\epsilon_n)= \log (\cosh (\epsilon_n))
    \label{emprical_risk}
\end{equation}
Here $\epsilon_n$ is the point-wise accuracy error and is defined as:  
\begin{equation}
    \epsilon_{n}=y_{n}-y\left(x_{n}\right)
    \label{eps1}
\end{equation}

For the uncertainty-focused part, $\ell(\cdot)$ is constructed via a pinball loss $\rho(\cdot)$ that is defined as: 
\begin{equation}
\rho(x_n, y_n, \tau)= \max({\tau}(y_{n}-{y}({x}_{n})),({\tau}-1)(y_{n}-{y}({x}_{n})))
\label{pinball_loss_lower}
\end{equation}
Here $\tau$ defines the desired quantile level to be covered. For learning an envelope that captures the expected amount of uncertainty, we define a lower $(\underline{\tau})$ and upper $(\overline{\tau})$ quantile level. We utilize TR set of $\alpha_0$-plane, $[\underline{y}^{\alpha_0}(x_{n}), \overline{y}^{\alpha_0}(x_{n})]$ as our lower and upper bound predictions and define the following loss:
\begin{equation}
    \ell\left(x_n, y_n,\underline{\tau}, \overline{\tau}\right) = \underline{\ell}_{\tau}^{\alpha_{0}}\left(x_n, y_n,\underline{\tau}\right) + \overline{\ell}_{\tau}^{\alpha_{0}}\left(x_n, y_n,\overline{\tau}\right) \label{eq:tilted} 
\end{equation}
with
\begin{equation}\label{eq:tilted_lower} 
\underline{\ell}_\tau^{\alpha_0}= \max(\underline{\tau}(y_{n}-\underline{y}^{\alpha_0}({x}_{n})),(\underline{\tau}-1)(y_{n}-\underline{y}^{\alpha_0}({x}_{n})))
\end{equation}
\begin{equation}\label{eq:tilted_upper}
    \overline{\ell}_\tau^{\alpha_0}=\max(\overline{\tau}(y_{n}-\overline{y}^{\alpha_0}({x}_{n})),(\overline{\tau}-1)(y_{n}-\overline{y}^{\alpha_0}({x}_{n})))
\end{equation}

To summarize, we define the following loss function $L$ for learning dual-focused GT2-FLS:
\begin{align}
 L=\frac{1}{N} \sum_{n=1}^{N}\left[L_{R}\left(\epsilon_{n}\right)+\ \ell\left(x_n, y_n,\underline{\tau}, \overline{\tau}\right)\right] \label{loss1}
\end{align}
Via the loss function, an (partially) independent learning of $\theta$ is feasible to capture uncertainty while achieving high accuracy.

\begin{algorithm} 
\caption{DL-based Dual-Focused GT2-FLS}
\begin{algorithmic}[1]
\label{alg:integration}
\State \textbf{Input:} $N$ training samples $(x_{n},y_{n})^{N}_{n=1}, \phi = [\underline{\tau}, \overline{\tau}]$
\State $K+1$, number of $\alpha$-planes 
\State $P$, number of rules
\State $mbs$, mini-batch size
\State \textbf{Output:} LP set ${\theta}$
\State Initialize ${\theta} = [{\theta}_{A}, {\theta}_{C}]$;
\State Perform tricks ${\theta}$; \algorithmiccomment{Sec. III.D}
\For{\textbf{each } $mbs$ in $N$} 
    \State $\mu^* \leftarrow \text{PMF}(x; {{\theta}}_{{AP}})$ \Comment{ Eq. \eqref{newpmf}} 
    \State $[\underline{\mu}^{\alpha_k}, \overline{\mu}^{\alpha_k}] \leftarrow \text{SMF}(\mu^*; {{\theta}}_{{AS}})$ 
    \Comment{ Eq. \eqref{muupper}}
    \State $[\underline{y}^{\alpha_0}, \overline{y}^{\alpha_0}, y] \leftarrow \text{Inference}(\underline{\mu}^{\alpha_k}, \overline{\mu}^{\alpha_k}; {\theta}_C)$ \Comment{Sec. II}
    \State Compute $L$ \Comment{Eq. \eqref{loss1}}
    \State Compute the gradient ${\partial L}/{\partial {\theta}}$
    \State Update ${\theta}$ via Adam optimizer
\EndFor

\State ${\theta}^* = {\arg \min }(L)$
\State \textbf{Return} $\theta = {\theta}^{*}$
\end{algorithmic}
\end{algorithm}

\begin{table*}
\centering
\caption{Testing Performance Analysis Over 20 Experiments}
\begin{threeparttable}
\begin{tabular}{cccccccl}
\hline
Dataset & $M\times N$ & Metric & Z-GT2-FLS & MJ-GT2-FLS & IT2-FLS-H & IT2-FLS-HS\\
\hline
White Wine & $11\times4898$ & $\#$LP & 192 & 247 & 225 & 280\\
     & & RMSE & \textbf{79.80($\pm$1.45)} & 81.44($\pm$2.05)  & 82.40($\pm$2.92) & 82.62($\pm$2.46)\\
     & & PICP & \textbf{97.73($\pm$0.62)} & 97.32(($\pm$1.14)  & 97.09($\pm$1.52) & 97.54($\pm$1.58)\\
     & & PINAW & 76.38($\pm$8.07) & 75.54($\pm$10.70)  & 74.39($\pm$11.40) & \textbf{74.35($\pm$9.12)}\\
\hline
Parkinson's  & $19\times5875$ & $\#$LP & 328 & 423 & 385 & 480\\
          Motor UPDRS & & RMSE & \textbf{60.42($\pm$4.14)} & 75.24($\pm$9.98) & 88.15($\pm$9.80) & 74.17($\pm$10.06)\\
          & & PICP & \textbf{98.95($\pm$0.66)} & 96.83($\pm$3.82)  & 91.60($\pm$4.31) & 98.20($\pm$2.88)\\
          & & PINAW & \textbf{142.53($\pm$29.25)} & 163.51($\pm$34.04)  & 165.02($\pm$36.49) & 171.68($\pm$39.58)\\
\hline
AIDS & $23\times2139$ & $\#$LP & 396 & 511 & 465 & 580\\
     & & RMSE & \textbf{69.74($\pm$2.56)} & 70.25($\pm$2.91)  & 72.30($\pm$2.75) & 71.94($\pm$3.52)\\
     & & PICP & 97.53($\pm$0.84) & 94.85($\pm$2.88) & 97.00($\pm$1.24) & \textbf{98.64($\pm$0.69)}\\
     & & PINAW & \textbf{160.35($\pm$18.49)} & 160.70($\pm$21.43)  & 185.57($\pm$14.15) & 210.76($\pm$19.73)\\
\hline
\end{tabular}
\begin{tablenotes}
\item (1) RMSE and PINAW values are scaled by 100. \item (2) Measures that are highlighted indicate the best ones.
\end{tablenotes}
\end{threeparttable}
\label{tab:dataset-results}
\end{table*}

\section{Performance Analysis} \label{performance}
In this section, we analyze the learning performance of the proposed Z-GT2-FLS compared to MJ-GT2-FLS and two IT2-FLSs on the high dimensional benchmark datasets, which are White Wine, Parkinson Motor UPDRS, and AIDS.

\subsection{Design of Experiments}
All datasets are preprocessed by Z-score normalization, and we split $70 \%$ of the data into the training set and $30 \%$ of the data into the test set. For each dataset, we trained all FLSs using identical hyperparameters, including mbs and learning rate, for 100 epochs, except for the Parkinson dataset, where we extended the training to 1000 epochs. The desired PI coverage was set as $99 \% (\varphi=$ $[0.005,0.995])$. The experiments were conducted within MATLAB ${ }^{\circledR}$ and repeated with 20 different initial seeds for statistical analysis. 

We configured the GT2-FLSs with 3 $\alpha$-planes while the IT2-FLSs, namely IT2-FLS-H and IT2-FLS-HS, are defined with IT2-FSs (Definition \ref{mendel}, in which $\mu_A(x, u) = 1$) and are learned via the DL-based approach presented in \cite{beke}. IT2-FLS-H is a simpler form of IT2-FLS-HS as $\underline{\sigma}_{p, m} = \overline{\sigma}_{p, m}, \forall p, m$. Thus, IT2-FLS-H has in total $3PM$ + $P(M+1)$ LPs while IT2-FLS-HS has $4PM$ + $P(M+1)$ LPs. In the experiments, we set $P=5$ for all FLSs. The number of LP (\#LP) of FLSs for the handled datasets are tabulated in Table \ref{tab:dataset-results}. It can be observed that the IT2-FLS-HS has the largest number of LPs.

\subsection{Performance Evaluation} 
We evaluated the performances via Root Mean Square Error (RMSE), PI Coverage Probability (PICP), and PI Normalized Averaged Width (PINAW) \cite{PINAW}. We anticipate training a T2-FLS that yields a low RMSE, i.e., high point-wise estimation, and attains a PICP close to 99\% with a low PINAW value, thereby indicating an HQ-PI as defined in \cite{pearce}.

\begin{figure}[hbpt]
\centerline{\includegraphics[scale=0.64]{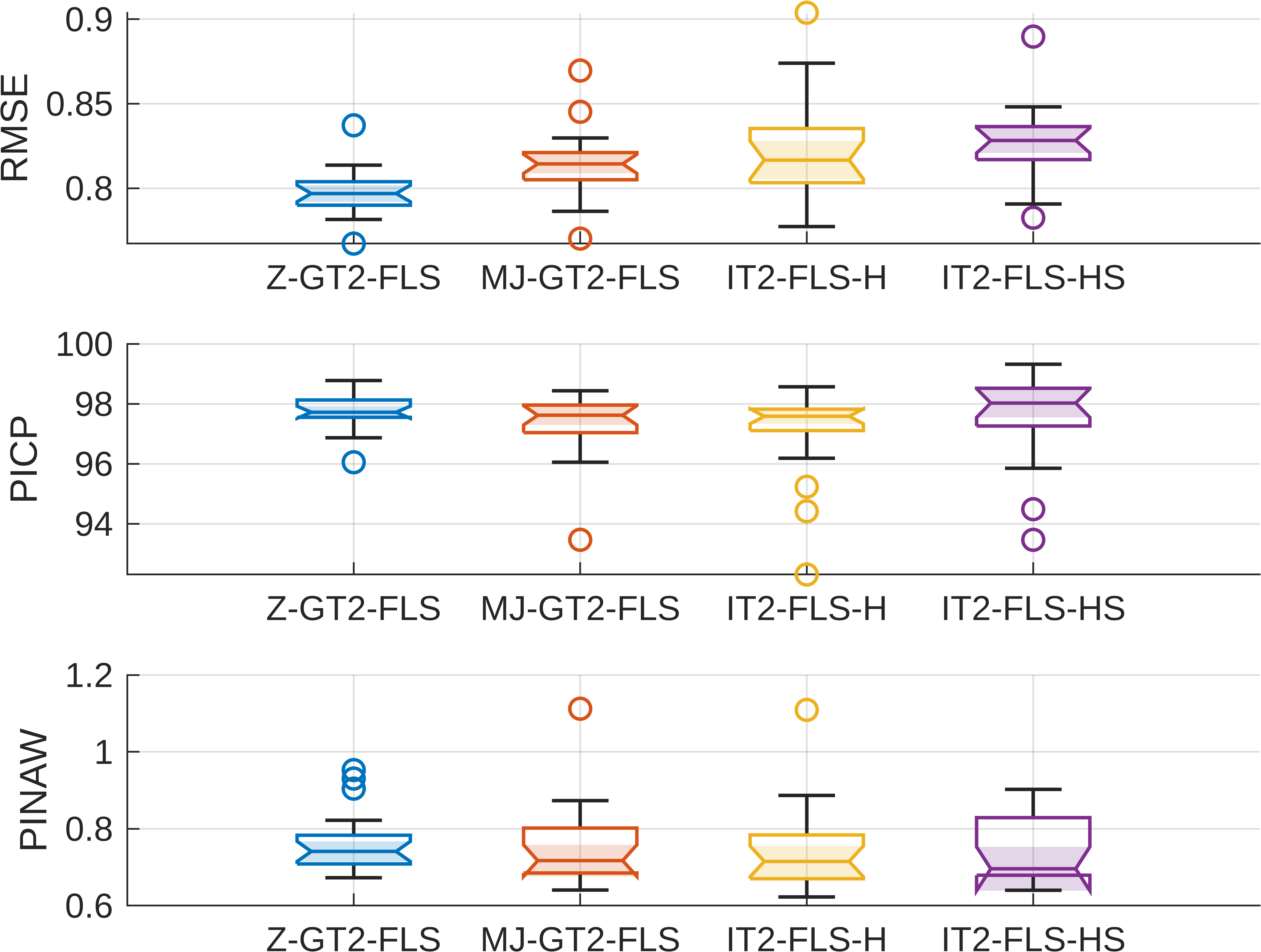}}
\caption{White Wine dataset: Notched box-and-whisker plots}
\label{fig1_wine}
\end{figure}

\begin{figure}[hbpt]
\centerline{\includegraphics[scale=0.63]{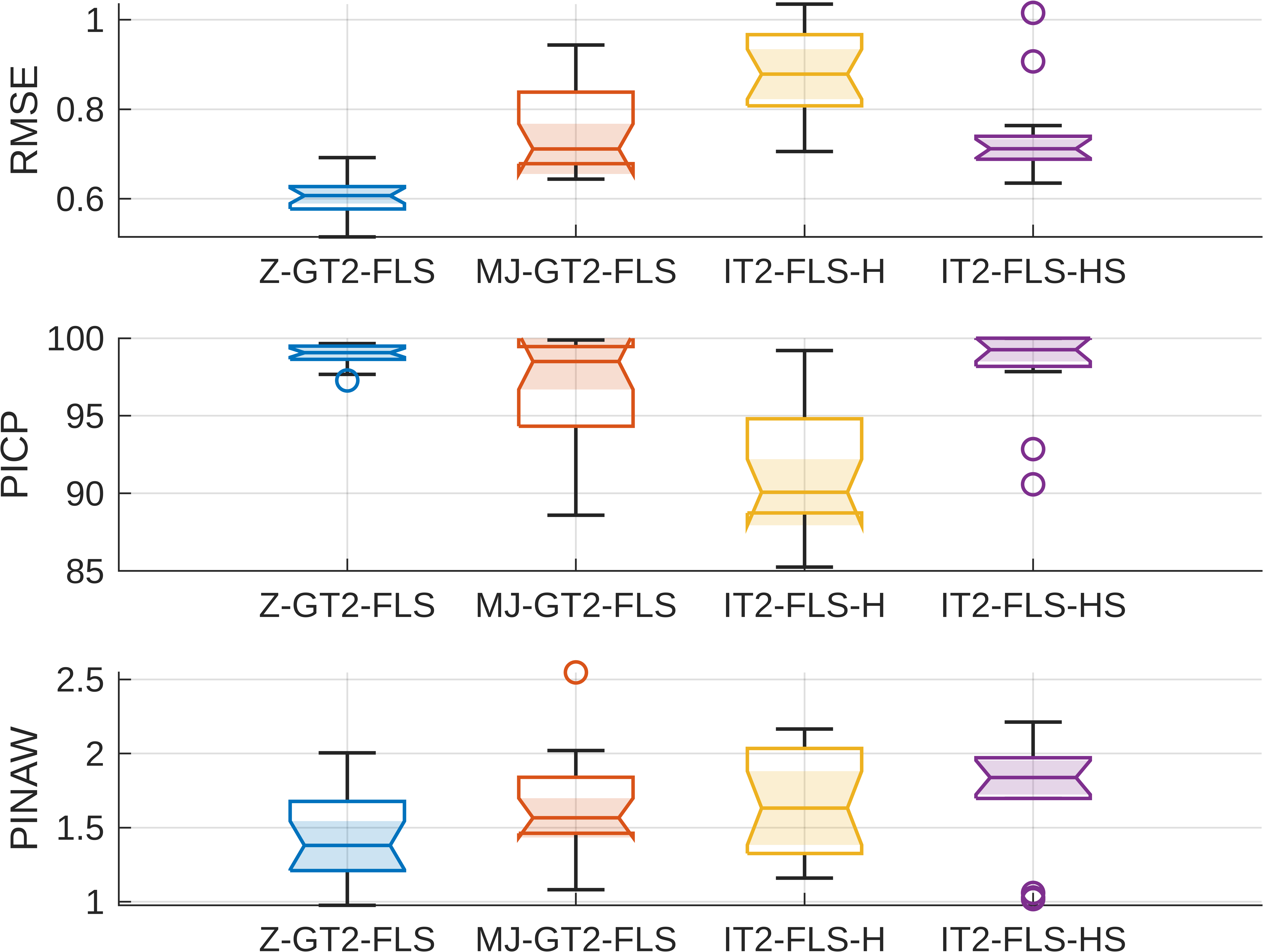}}
\caption{Parkinson dataset: Notched box-and-whisker plots}
\label{fig1_parkinson}
\end{figure}

\begin{figure}[hbpt]
\centerline{\includegraphics[scale=0.63]{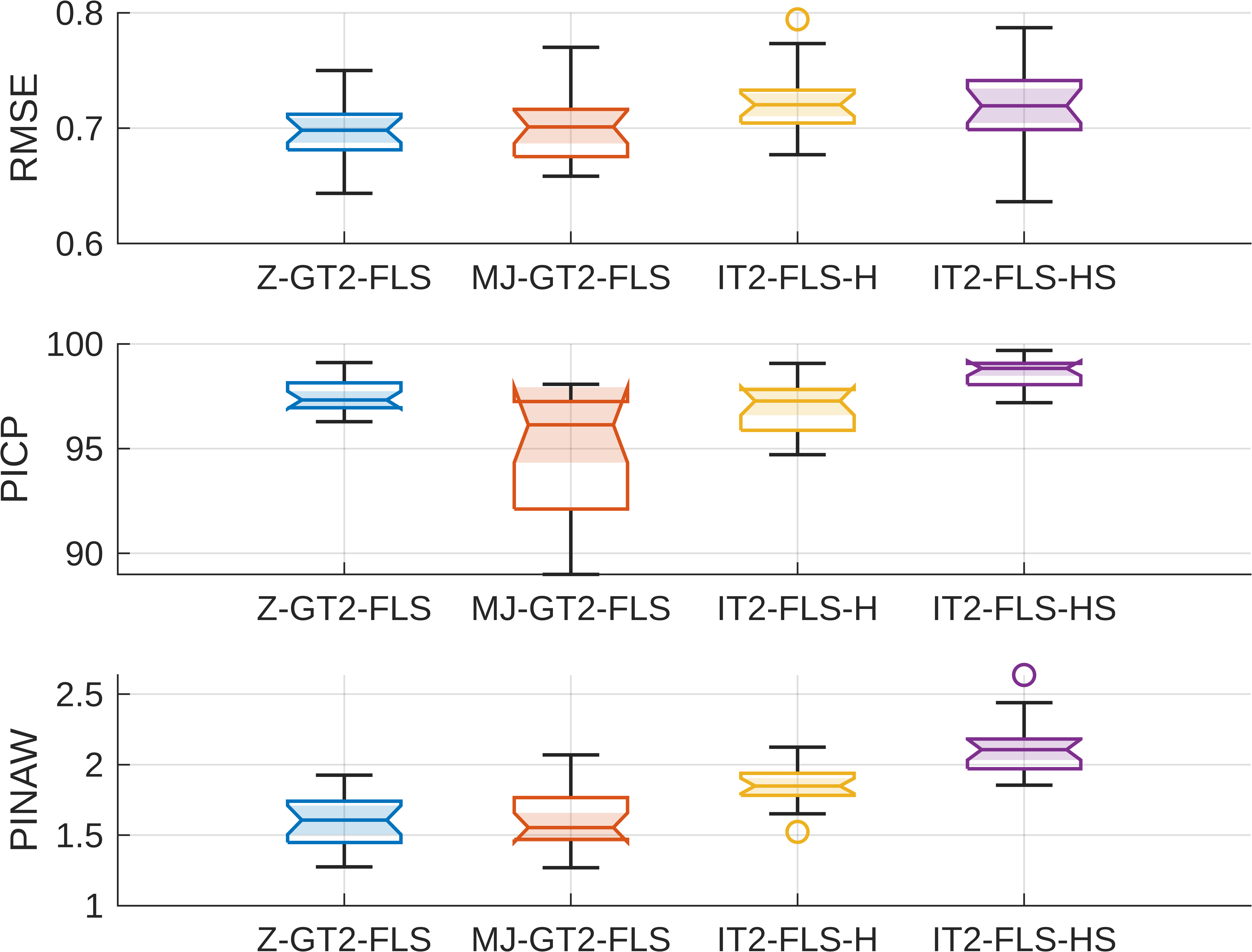}}
\caption{AIDS dataset: Notched box-and-whisker plots}\label{fig1_aids}
\end{figure}

Table \ref{tab:dataset-results} provides the mean RMSE, PICP, and PINAW alongside their $\pm1$ standard error values while Fig. \ref{fig1_wine}---Fig.\ref{fig1_aids} presents the notched box and whisker plots showing median (central mark), $25^\text{{th}}$/$75^\text{{th}}$ percentiles (left and right edges of box) which defines the Inter Quartile Range (IQR), whiskers (line), and outliers (circles). Observe that:
\begin{itemize}
    \item For the White Wine dataset, the Z-GT2-FLS resulted in the best RMSE performance with a statistically significant difference, due to the absence of notches overlapping, as shown in Fig. \ref{fig1_wine}. It can also be stated that the width of IQR is smaller which also aligns with the standard errors of the measures presented in Table \ref{tab:dataset-results}, suggesting lower variability for different initial seeds. 
    \item For the Parkinson dataset, The Z-GT2-FLS model demonstrates superior performance with the lowest RMSE with a statistically significant difference. It also has the highest PICP with a narrow IQR and the lowest mean PINAW, indicating the generation HQ-PIs. IT2-FLS-HS also offers good coverage but at the cost of wider PIs.
    \item For the Aids dataset, the point-wise predictions of all FLSs are of similar performance outcomes, yet the Z-GT2-FLS has a better mean performance in RMSE. On the other hand, as shown in Fig. \ref{fig1_aids}, the IT2-FLS-HS has the best PICP, yet its PINAW is significantly larger than its counterpart, indicating a poor PI. On the other hand, Z-GT2-FLS has a similar PI performance with a much narrower PI band (i.e. low PINAW) and thus an HQ-PI.

\end{itemize}

To sum up, Z-GT2-FLS generally offers the best accuracy across all datasets. It maintains good to excellent coverage (high PICP) with tight PI widths (low PINAW), making it a consistently robust performer.

\section{Conclusion and Future Work} \label{Conc}
In this paper, we constructed and learned GT2-FLSs, based on Zadeh's definition, with design flexibility on SMF, unlike its MJ counterpart. We introduced a DL framework that does not violate the constraints of the Z-GT2-FLS via parameterization tricks while also taking care of the curse of dimensionality. The presented results have shown that the Z-GT2-FLS demonstrates a high degree of predictive accuracy and reliability with HQ-PIs on high-dimensional datasets although it has fewer LPs when compared to IT2-FLSs and MJ-GT2-FLS.

As for our future work, we plan to develop a learning framework for GT2-FLSs for merely UQ.



\section*{Acknowledgment}
The authors acknowledge using ChatGPT to refine the grammar and enhance the English language expressions.

\bibliographystyle{IEEEtran}
\bibliography{IEEEabvr,cites}

\end{document}